 \documentclass[wcp]{jmlr}% former name JMLR W\&CP

\RequirePackage{graphicx}
 \usepackage{booktabs}
 \usepackage{multirow}
\usepackage{longtable}% for long tables
\makeatletter
\def\set@curr@file#1{\def\@curr@file{#1}} %temp workaround for 2019 latex release
\makeatother
\usepackage[load-configurations=version-1]{siunitx} % newer version

\theorembodyfont{\upshape}
\theoremheaderfont{\scshape}
\theorempostheader{:}
\theoremsep{\newline}

\title[Psychiatry Reliability Audit]{Reliability Auditing for Downstream LLM tasks in Psychiatry: LLM-Generated Hospitalization Risk Scores}

\author{\Name{Shevya Panda} \and \Name{Shinjini Bose} \and \Name{Ananya Joshi}
\addr Johns Hopkins University\\ Baltimore, USA}

\begin{document}

\maketitle

\begin{abstract}
  Large language models (LLMs) are increasingly utilized in clinical reasoning and risk assessment. However, their interpretive reliability in critical and indeterminate domains such as psychiatry remains unclear. Prior work has identified algorithmic biases and prompt sensitivity in these systems, raising concerns about how contextual information may influence model outputs, but there remains no systematic way to assess these, especially in the psychiatric domain. We propose an approach for reliability auditing downstream LLM tasks by structuring evaluation around the impact of prompt design and the inclusion of medically insignificant inputs on predicted hospitalization risk scores, which is often the first downstream AI clinical-decision-making task. In our audit, a cohort of synthetic patient profiles (n = 50) is generated, each consisting of 15 clinically relevant features and up to 50 non-clinically relevant features, across four prompt reframings (neutral, logical, human impact, clinical judgment). We audit four LLMs (Gemini 2.5 Flash, LLaMa 3.3 70b, Claude Sonnet 4.6, GPT-4o mini), and our results show that including medically insignificant variables resulted in a statistically significant increase in the absolute mean predicted hospitalization risk and output variability across all models and prompts, indicating reduced predictive stability as contextual noise increased. Clinically insignificant features had an effect on instability across many model-prompt conditions (Simple $\beta$1, $p < 0.05$), and prompt variations independently affected the trajectory of instability in a model-dependent manner. These findings quantify how LLM-based psychiatric risk assessments are sensitive to non-clinical information, highlighting the need for systematic evaluations of attributional stability and uncertainty behavior like this before clinical deployments.

\end{abstract}

\section{Introduction}

 Recent breakthroughs and ongoing developments of Large Language Models (LLMs) have enabled their integration into healthcare systems, and many proposed healthcare decision-making tools \citep{Jiang2021, Busch2025, Jiang2023}. In particular, new research in clinical psychiatry indicates that LLMs can serve as helpful tools for supporting complex decision-making, spanning diagnosis, treatment, and prognosis to enhance patient care and physician workflows \citep{Perlis2024}. Still, the downstream sensitivity of these models is understudied in highly subjective healthcare domains like psychiatric risk assessment of psychiatric disorders, where context and language nuances carry diagnostic weight and the complexity of medical language / incomplete or inconsistent information available to models might be particularly impactful \citep{Li2025}.

 Thus, there is a need to deeply evaluate the reliability of these models. Many psychiatric AI applications commercially emphasize the training data used (e.g. restriction fine tuning or training only to clinical notes). However, models have bias from their training data and there is also instability in how models internally weigh information \citep{ibm_llm, Ayoub2024, Omiye2023}. Thus, LLMs can display inconsistent attribution patterns and assign varied importance to similar features across different prompts or contexts and may also encode and reproduce judgments implicit in the language they are trained on \citep{Jiang2021, Ayoub2024, garcia2024moral}. In healthcare applications, where stable and interpretable reasoning is prioritized, increased predictive instability may suggest unreliable reasoning processes rather than baseline statistical noise impacting downstream AI application diagnostic reasoning, patient risk assessments, and care recommendations across any field of medicine dependent on demographics or behavioral data \citep{garcia2024moral, HadarShoval2024}. In addition, altering the input data available to LLMs in clinical risk assessment may change the model's interpretation, altering how they classify risk. Therefore, adding insignificant features may alter the model's rationale and risk calibration, and semantically meaningless inputs cause the model to redistribute importance among features \citep{Johnson2023, Ghorbani2019}. Clinical risk assessments can also shift based on when inputs have no medical relevance, making this a systemic concern that extends well beyond any single specialty. These limitations contribute to growing mistrust of LLM utilization in healthcare among both physicians and patients \citep{Busch2025}. Recognizing and mitigating these inconsistencies is therefore essential for ensuring that model outputs remain transparent, equitable, and clinically valid. Therefore, the foundation of such biases must be surfaced quickly.

\begin{figure}
    \centering
    \includegraphics[width=1\linewidth]{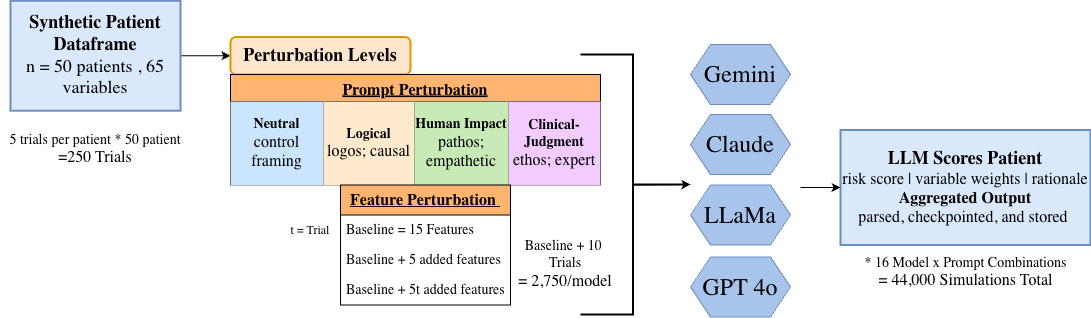}
    \caption{Experimental pipeline for evaluating LLM sensitivity to clinically insignificant features. A synthetic cohort of 50 patient profiles was evaluated across four large language models and four prompt styles, with each patient profile systematically amended from a 15-feature clinical baseline to a maximum of 65 features through incremental addition of clinically insignificant variables in batches of five, creating 11 feature configurations per patient. Each model, prompt, patient configuration combination is repeated 5 times for stochastic variability, totaling 44,000 simulations.}
    \label{fig:placeholder}
\end{figure}

Accordingly, this study aims to examine the extent to which LLMs are affected by clinically insignificant variables in generating psychiatric risk assessment scores.  Psychiatric disorders are already difficult to diagnose and determine next step care, primarily hospitalization as it represents a consequential decision clinicians must make, and there is an increasing application of machine learning models to improve diagnostic precision \citep{Luciano2025, Narimani2025, sekhon2023mood}. If LLMs systematically devalue or stigmatize certain groups based on medically insignificant information such as behaviors and extraneous genetic traits, it can worsen patient outcomes \citep{Cronj2023}. Understanding these mechanisms is crucial before clinical deployment. Therefore, the question of interest is to what extent the inclusion of medically insignificant variables alters the interpretive and predictive stability of the LLM’s predicted risk for hospitalization. 

\subsection*{Generalizable Insights about Machine Learning in the Context of Healthcare}

\begin{enumerate}
    \item Structured audits can surface failure modes that would otherwise be missed, especially in psychiatric settings where outputs are sensitive to subtle changes. This is particularly important as deployed systems may rely on interchangeable or frequently updated LLMs.
    \item Prompt engineering is an often-overlooked variable in LLM-based clinical diagnostic support. This work demonstrates that rhetorical style, independent of clinical content, can systematically shift model outputs. This emphasizes that prompt design must be a controlled methodological choice in any AI-based healthcare pipeline.
    \item Sensitivity to added noise can be used to test reliability. Adding irrelevant features and tracking output stability provides a general way to audit LLM performance.
    
    \item We provide a reusable audit framework. By adding irrelevant information and varying prompts, this approach helps identify when models fail to prioritize clinically relevant features before deployment.
\end{enumerate}

\section{Related Work}
We study reliability by altering prompting strategies and including additional features in patient profiles. Prompt engineering plays a critical role in how LLMs interpret and respond to input data, and forms the basis of our reliability analysis. Minute changes in wording or structure can lead to significant differences in responses, highlighting the sensitivity of models to input design \citep{Sonoda2024}. Previous studies suggest that structured prompting is an effective approach to improve diagnostic performance \citep{Dymm2026}, and we use this approach to reduce variability and promote reliable output responses. 

Several works also demonstrate model fragility and sensitivity to irrelevant features\citep{Ghorbani2019, Johnson2023, khandelwal_llm_confusion}. Due to the hidden logistics of black box models, it is difficult to attribute which features are given importance in the so-called “reasoning process” of LLMs. The interpretation of features by a model may alter, even if the overall prediction remains the same. LLMs have also been shown to inherently encode human-like value systems that influence their ethical and interpretive reasoning, which carries implications in their integration into medical diagnostic systems \citep{Bai2025, HadarShoval2024, Jiang2021, Ayoub2024, garcia2024moral}.In terms of clinical decision-making, existing literature indicates that LLMs demonstrate robust diagnostic accuracy on real medical data \citep{Yao2025}. These findings are further corroborated in comparative analyses of varied LLM architectures, where top-performing models are evaluated for diagnostic ability and show high performance rates \citep{Sarvari2025, Dinc2025}. Still, in a clinical diagnostic context, inconsistent or conflicting outputs, even if the model is proportionally correct, would undermine the reliability of practical utilization of LLMs in healthcare. Accordingly, we interpret the main findings of the current body of research to indicate that the addition of clinically non-salient features may alter the model’s risk calibration and redistribute importance among features.

\section{Problem Formulation}

The primary hypothesis of this study is that LLMs exhibit measurable predictive instability when presented with normatively insignificant patient information, reflecting sensitivity to algorithmically encoded biases. Specifically, we test whether the inclusion of medically insignificant variables in a patient profile , and the prompting strategy, produces systematic changes in reported hospitalization risk scores, despite these variables having limited to no relationship. We strategically grounded each prompt under rhetorical frameworks:logos, ethos, pathos, along with a neutral prompt for baseline comparison. These modes have been recognized as fundamental communication and persuasive measures, and clinicians often map onto one of these rhetoric's when reasoning patient profiles. 

Formally, let \(k \in \mathcal{K}\) denote a patient,  \(p \in \{\text{neutral, logos, ethos, pathos}\}\) one of the prompting strategies, , \(c_t\) a feature configuration, and \(m\) denote one of the four LLMs to audit. The baseline configuration \(c_0\) contains only clinically relevant patient features. Each non-baseline configuration \(c_t\), for \(t=1,\dots,T\), augments \(c_0\) with additional medically irrelevant variables. In our setup, configuration \(c_t\) contains \(5t\) added irrelevant features. For each model \(m\), prompt \(p\), patient \(k\), configuration \(c_t\), and stochastic repetition \(s \in \{1,\dots,S\}\), let
\[
r_m(p,c_t,k,s)
\]
denote the hospitalization risk score returned by the model, where \(S=5\) repeated generations are used to capture within-condition variability. We define the mean predicted risk under a given condition as
\begin{align}
\bar r_m(p,c_t,k)
=
\frac{1}{R}\sum_{s=1}^{S} r_m(p,c_t,k,s).
\end{align}

To quantify instability, we compare each non-baseline configuration to the baseline prediction for the same patient and prompt. Specifically, we define predictive instability as the absolute deviation from baseline:
\begin{align}
\Delta_m(p,c_t,k)
=
\left|
\bar r_m(p,c_t,k)
-
\bar r_m(p,c_0,k)
\right|.
\end{align}

Thus, \(\Delta_m(p,c_t,k)\) measures how much the model's average predicted risk changes when irrelevant variables are added to an otherwise identical patient profile. Larger values of \(\Delta_m(p,c_t,k)\) indicate greater sensitivity to irrelevant information. 

% To summarize instability across patients and non-baseline configurations for a given model and prompting strategy, we define
% \begin{align}
% \mu_{m,p}
% =
% \frac{1}{|\mathcal{K}|\,T}
% \sum_{k \in \mathcal{K}}
% \sum_{t=1}^{T}
% \Delta_m(p,c_t,k).
% \end{align}

\subsection{Statistical Evaluation}
To evaluate factors contributing to instability, we use mixed-effects models. The first set of tests evaluates whether predictive instability increases as medically irrelevant features are added, separately within each prompting strategy. For each LLM and each prompting strategy \(p\), we fit
\[\textbf{Basic: }
\Delta_m(p,c_t,k)
=
\alpha_0^{(p)} + \alpha_1^{(p)} t + u_k + \epsilon_{ptk},
\]
and test
\[
\begin{aligned}
H_{1,\text{m, p}}^{(0)} &: \alpha_1^{(\text{m, p})} = 0 
&\quad&
H_{1,\text{m, p}}^{(A)} &: \alpha_1^{(\text{m, p})} > 0 \\
\end{aligned}
\]
Rejection of \(H_{1,p}^{(0)}\) indicates that predictive instability increases as medically irrelevant features are added under prompting strategy \(p\). These tests assess feature-induced instability within each prompt independently and do not compare across prompting strategies.
The remaining three hypotheses come from the fuller model fit separately for each LLM:
\[\textbf{Complex: }
\Delta_m(p,c_t,k)
=
\beta_0
+
\beta_1 t
+
\sum_{j \in \{\text{logos, ethos, pathos}\}}
\Big(
\beta_{2j}
+
\beta_{3j} \, t
\Big)
\mathbf{1}\{p=j\}
+
u_k
+
\epsilon_{ptk}.
\]

\(\beta_0\) represents the expected instability for the default prompting strategy at the lowest non-baseline configuration, \(\beta_1\) captures the change in instability with feature accumulation under the default prompting strategy, \(\beta_{2}\) captures the difference in baseline instability between prompting strategy \(j\) and the default strategy, and \(\beta_{3}\) captures the difference in slope with respect to feature accumulation between the prompting strategy \(j\) and the default strategy.

where neutral prompting is the reference category. First, we test whether instability increases with irrelevant-feature accumulation for the neutral prompting strategy:
\[
\begin{aligned}
H_{2, \text{m, p}}^{(0)} &: \beta_1^{(\text{m, p})} = 0 &\quad&
H_{2, \text{m, p}}^{(A)} &: \beta_1{(\text{m, p})} > 0
\end{aligned}
\]

Second, we test whether each prompting strategy differs from neutral in baseline instability:
\[
\begin{aligned}
H_{3, \text{m, p}}^{(0)} &: \beta_2^{(\text{m, p})} = 0 &\quad&
H_{3, \text{m, p}}^{(A)} &: \beta_2{(\text{m, p})} \neq 0
\end{aligned}
\]

Third, we test whether each prompting strategy differs from neutral in how instability changes with feature accumulation:
\[
\begin{aligned}
H_{4, \text{m, p}}^{(0)} &: \beta_3^{(\text{m, p})} = 0 &\quad&
H_{4, \text{m, p}}^{(A)} &: \beta_3{(\text{m, p})} \neq 0
\end{aligned}
\]

Per model, the rejection of \(H_4^{(0)}\) indicates that at least one prompting strategy differs from the default strategy in how instability changes as irrelevant information accumulates. To account for multiple hypothesis testing across these hypotheses, we control the family-wise error rate using a Bonferroni-adjusted significance threshold \textit{per model}:
\[
\alpha_{\mathrm{corr}} = \frac{0.05}{11} \approx 0.004.
\]

\section{Evaluation Framework Overview
}

As shown in Fig. \ref{fig:placeholder}, we develop a fully synthetic framework to assess predictive instability in LLMs in reasoning for hospitalization risk score. Models are subject to systemized perturbations of inputs and prompts. This framework is a diagnostic benchmark and we aim to quantify the sensitivity of LLM outputs to changes that should be prescriptively insignificant under a structured risk assessment procedure.  This cross design enables systematic comparison of outputs while controlling for the confounding effects of model choice, prompt variant, and patient feature configuration, to support generalizations regarding the predictive stability of LLM-generated risk scores. Detailed methods are available in the supplementary materials. This setup can expose whether sensitivity to clinically insignificant features leads to unstable outputs, offering a window into how LLMs may encode bias or values in practice.

\subsection{Synthetic Data Generation}

To evaluate potential internal biases and instability in LLMs when assessing psychiatric risk, we created a synthetic dataset designed to simulate patient archetypes. The dataset was generated using conditional probabilities across 15 baseline actionable diagnostic features and traits to approximate demographic and clinical characteristics of a general population while allowing controlled variation across socially sensitive variables (i.e., gender identity, sexual orientation, or political affiliation, covered in Tabs. \ref{tab:baseline_cont} and \ref{tab:baseline_cat}. Using conditional probabilities prevented a disproportionate number of nonsensical profiles (i.e. age 19 and retired). 

The baseline dataset was then augmented with 50 additional variables that very likely or only minimally direct medical relevance (i.e., eye color, likes chocolate, pet preference, recycling habits, drinks coffee, or annual charity donation) (see Appendix for full list), thus clarifying the influence of relatively clinically insignificant features on LLM-generated hospitalization risk scores.

% Continuous variables
\begin{table}[t]
\centering
\caption{Continuous baseline characteristics of the synthetic patient cohort ($n=50$).}
\label{tab:baseline_cont}
\small
\setlength{\tabcolsep}{6pt}
\renewcommand{\arraystretch}{1.1}
\begin{tabular}{@{}lccc@{}}
\toprule
\textbf{Variable} & \textbf{Mean (SD)} & \textbf{Median [IQR]} & \textbf{Range} \\
\midrule
Age                   & 40.7 (17.7) & 37.5 [25.2--52.8] & 18--90 \\
BMI                   & 26.7 (5.8)  & 26.1 [23.2--29.7] & 13--47 \\
Weekly Alcohol Intake (Drinks) & 3.2 (2.0)   & 3.0 [2.0--4.0]    & 0--7 \\
Hours of Sleep        & 7.0 (1.6)   & 7.0 [5.7--8.0]    & 3--10 \\
\bottomrule
\end{tabular}
\end{table}

\begin{table}[!h]
\centering
\caption{Categorical baseline characteristics of the synthetic patient cohort ($n=50$). For each categorical variable, the table reports the number of categories and the most frequent category.}
\label{tab:baseline_cat}
\small
\setlength{\tabcolsep}{6pt}
\renewcommand{\arraystretch}{1.1}
\begin{tabular}{@{}lccc@{}}
\toprule
\textbf{Variable} & \textbf{Categories} & \textbf{Modal category} & \textbf{Percent} \\
\midrule
Family Health History & 5 & Asthma & 22\% \\
Current Diagnosis     & 4 & Bipolar & 24\% \\
Race                  & 6 & Asian & 18\% \\
Gender                & 4 & Non-binary & 36\% \\
Sexual Orientation    & 5 & Heterosexual & 74\% \\
Smoking               & 2 & No & 74\% \\
HAM-D                 & 4 & 0--6: Normal / No Depression & 44\% \\
Been Sad or Fatigued  & 2 & No & 66\% \\
Loss of Interest      & 2 & Yes & 58\% \\
Employment            & 3 & Employed & 66\% \\
Socioeconomic Status  & 3 & Middle & 56\% \\
\bottomrule
\end{tabular}
\end{table}

\subsection{Perturbation Designs}

\paragraph{Prompt Perturbation Design}: Each of the models is evaluated under four prompt variants, as shown in Table \ref{ref:table3}, including the default. The logical prompt relies on explicit causal impacts, depicting logos. Human impact primarily relies on empathetic human understanding, highlighting pathos. Finally, the clinical judgment prompt framing takes on a specialist role, taking into account hollistic interpretations of patient data for ethos.

\paragraph{Feature Perturbation Design}: Including clinically insignificant features via a perturbation design allows us to systemically quantify how additional inputs influence risk scores. Our design iteratively introduces randomly selected clinically insignificant features in batches of 5 randomly-selected features each.  This is a more robust feature design because it lets us observe how the \textit{added number} features on average change risk scores, rather than just whether the effect exists for a single feature (highlighting the pattern of the effect itself). For each of the 50 patient profiles, a series of 11 datasets, comprising one baseline and 10 incrementally augmented feature sets.

\begin{table}[t]
\centering
\caption{Prompt variants used across all model evaluations. Prompt styles were held constant within conditions across all model--perturbation combinations. Neutral serves as the reference condition.}
\setlength{\tabcolsep}{4pt}
\small
\begin{tabular}{>{\raggedright\arraybackslash}p{2cm} >{\raggedright\arraybackslash}p{1.8cm} >{\raggedright\arraybackslash}p{3cm} >{\raggedright\arraybackslash}p{4cm} >{\raggedright\arraybackslash}p{3.5cm}}
\toprule
\textbf{Prompt Style} & \textbf{Rhetorical Mode} & \textbf{Clinical Analogue} & \textbf{Key Instruction Language} & \textbf{Calculation Constraint} \\
\midrule

Neutral
  & None (Control)
  & Generic clinical encounter
  & ``Assign a numerical risk score''
  & Exclude baseline risk \\[6pt]

Logical
  & Logos
  & Evidence-based medical specialist, causal reasoning
  & ``Formal, logic-based approach, explicitly reflecting causal relationships''
  & No baseline or population-level risk; contributions must sum exactly \\[6pt]

Human Impact
  & Pathos
  & Patient advocate, empathetic, social medicine clinician
  & ``Personal consequences\ldots greatest concern for hospitalization risk''
  & Exclude baseline risk; emphasize vulnerability \\[6pt]

Clinical Judgment
  & Ethos
  & Holistic, authority-driven, attending physician/specialist
  & ``Acting as a clinical expert\ldots professional''
  & Exclude baseline risk; impacts must sum exactly \\

\bottomrule
\end{tabular}
\flushleft
\label{ref:table3}

\end{table}

We report output audits using four models via the OpenRouter API: Gemini 2.5 Flash, LLaMa 3.3 70b instruct, Claude Sonnet 4.6, and GPT-4o mini. The primary measure of evaluation was predictive instability, characterized by the absolute change in mean risk score relative to the baseline condition consisting of 15 clinically relevant variables. Each configuration was repeated five times to account for statistical randomness and then averaged to produce a stable estimate of model output.

\section{Results}

% Preamble requirements:
% \usepackage{booktabs}
% \usepackage{array}
% \usepackage{multirow}
% \usepackage{siunitx} % for scientific notation formatting

\begin{table}[h!]
\centering
\setlength{\tabcolsep}{4pt}
\caption{Linear regression model comparison $p$-values across models and prompt styles. Simple $\alpha_1$ = effect of added irrelevant variables alone; Complex $\beta_1$ = combined model intercept; Complex $\beta_2$ = prompt main effect; Complex $\beta_3$ = prompt $\times$ added variables interaction. --- = not applicable (reference condition).}
\label{tab:pvalues}
\small
\begin{tabular}{>{\raggedright\arraybackslash}p{2.8cm} >{\raggedright\arraybackslash}p{2cm} >{\raggedright\arraybackslash}p{2cm} >{\raggedright\arraybackslash}p{2.2cm} >{\raggedright\arraybackslash}p{2.2cm} >{\raggedright\arraybackslash}p{2.2cm}}
\toprule
\textbf{Model} & \textbf{Prompt} & \textbf{Simple $\alpha_1$} & \textbf{Complex $\beta_1$} & \textbf{Complex $\beta_2$} & \textbf{Complex $\beta_3$} \\
\midrule
\multirow{4}{*}{Claude Sonnet 4.6}
  & Neutral            & $<$0.001*** & 0.008** & ---         & ---         \\
  & Clinical Judgement & 0.012*      & 0.008** & 0.094       & 0.178       \\
  & Human Impact       & 0.603       & 0.008** & 0.011*      & 0.058       \\
  & Logical            & 0.999       & 0.008** & $<$0.001*** & $<$0.001*** \\
\midrule
\multirow{4}{*}{Gemini 2.5 Flash}
  & Neutral            & 0.057       & 0.191   & ---         & ---         \\
  & Clinical Judgement & 0.954       & 0.191   & 0.013*      & 0.194       \\
  & Human Impact       & 0.005**     & 0.191   & 0.950       & 0.577       \\
  & Logical            & 0.673       & 0.191   & $<$0.001*** & 0.317       \\
\midrule
\multirow{4}{*}{LLaMA 3.3 70B}
  & Neutral            & $<$0.001*** & 0.009** & ---         & ---         \\
  & Clinical Judgement & 0.002 $\diamond$     & 0.009** & 0.631       & 0.791       \\
  & Human Impact       & 0.183       & 0.009** & 0.300       & 0.221       \\
  & Logical            & 0.001**     & 0.009** & 0.042*      & 0.860       \\
\midrule
\multirow{4}{*}{GPT-4o mini}
  & Neutral            & $<$0.001*** & $<$0.001*** & ---    & ---         \\
  & Clinical Judgement & $<$0.001*** & $<$0.001*** & 0.858  & 0.054       \\
  & Human Impact       & $<$0.001*** & $<$0.001*** & 0.086  & 0.571       \\
  & Logical            & $<$0.001*** & $<$0.001*** & 0.039* & 0.001 $\diamond$     \\
\bottomrule
\end{tabular}
\footnotesize{$p$-values reported from linear regression. --- = reference category (neutral prompt; no prompt contrast applicable). *** $p < 0.001$; $\diamond$  $p < 0.004$; ** $p < 0.01$; * $p < 0.05$.}
\label{tab:REML}
\end{table}

Outputs from the mixed effects model is shown in Tab. \ref{tab:REML}. We evaluate our hypotheses using the mixed effects model across four terms: the effect of added variables alone (Simple $\alpha_1$), the combined model intercept (Complex $\beta_1$), the prompt main effect (Complex $\beta_2$), and the prompt × added variables interaction (Complex $\beta_3$).

We reject the null for GPT-4o mini, where all prompt conditions are significant under both Simple $\beta$1 and Complex $\beta_1$ ($p < 0.001$ throughout), making it the only model where the null is rejected in every condition tested. For LLaMA, we reject under neutral, logical, and clinical judgment ($p < 0.01$) but fail to reject for human impact ($p = 0.183$), with no significant interaction terms suggesting prompt framing shifts instability level without moderating accumulation. For Claude, we reject under neutral ($p < 0.001$) and clinical judgment ($p = 0.012$) but fail to reject for human impact and logical under Simple $\beta$1, the logical interaction term is the most significant result in the entire dataset ($p < 0.001$), indicating logical prompting changes how Claude responds to noise rather than simply shifting its baseline. Gemini is the only model where we fail to reject the null under Simple $\alpha_1$ for the majority of prompt conditions, with the combined intercept also non-significant ($p = 0.191$), suggesting noise accumulation alone does not reliably drive instability in this model --- its most notable finding is instead the logical prompt main effect ($p < 0.001$), suggesting that logical framing produces a large level shift independent of noise.

\subsection{Prompt Sensitivity} 

Alongside added features, prompt framing significantly affects the trajectory of instability in a manner that varies across models (Figure~\ref{fig:means}). For LLaMA, the logical prompt main effect was significant (Complex $\beta_2$, $p = 0.042$), with no significant interaction terms, suggesting prompt framing shifts instability level without moderating accumulation. Gemini showed the largest prompt main effect under logical framing (Complex $\beta_2$, $p < 0.01$), though the interaction was not significant (Complex $\beta_3$, $p = 0.317$), confirming the effect is level-driven rather than cumulative. For Claude, the logical interaction term was the most significant result in the dataset (Complex $\beta_3$, $p < 0.01$), indicating logical framing alters how Claude responds to noise rather than shifting its baseline. GPT-4o mini was the only model where logical framing significantly affected both level and trajectory (Complex $\beta_2$, $p = 0.039$; Complex $\beta_3$, $p = 0.001$). Clinical judgment was not significant across most model and prompt combinations, and was the most stable prompt style across all models (Figure~\ref{fig:alt}).

Across all models, the addition of the first five irrelevant variables produces the largest single shift in mean instability (Figure~\ref{fig:means}), suggesting models are more sensitive to the initial introduction of irrelevant information than to its continued accumulation. When looking at the confidence intervals, there are a few key patterns to note between models. Claude Sonnet has the widest CI in comparison to models, particularly at the first perturbation from baseline, showing that some patients are dramatically destabilized while other are not as affected (Figure \ref{fig:means} mean instability CI). GPT-4o mini has confidence intervals that progressively get wider with an increase in non-clinical variables, meaning that the uncertainty increases over perturbations. LLaMa has the narrowest CIs overall, followed by Gemini,  suggesting that these models have more consistent responses to noise regardless of prompt variation. 

\subsection{Added Features as a Driver of Instability}

% \begin{figure}[h]
%     \centering
%     \includegraphics[width=\linewidth]{image (8).png}
%     \caption{Distribution of absolute mean risk score instability across perturbation levels and prompt styles for each model. Box plots represent patient-level deviation from baseline across five repeated trials per condition. }
%     \label{fig:means}
% \end{figure}

\begin{figure}[h]
    \centering
    \includegraphics[width=\linewidth]{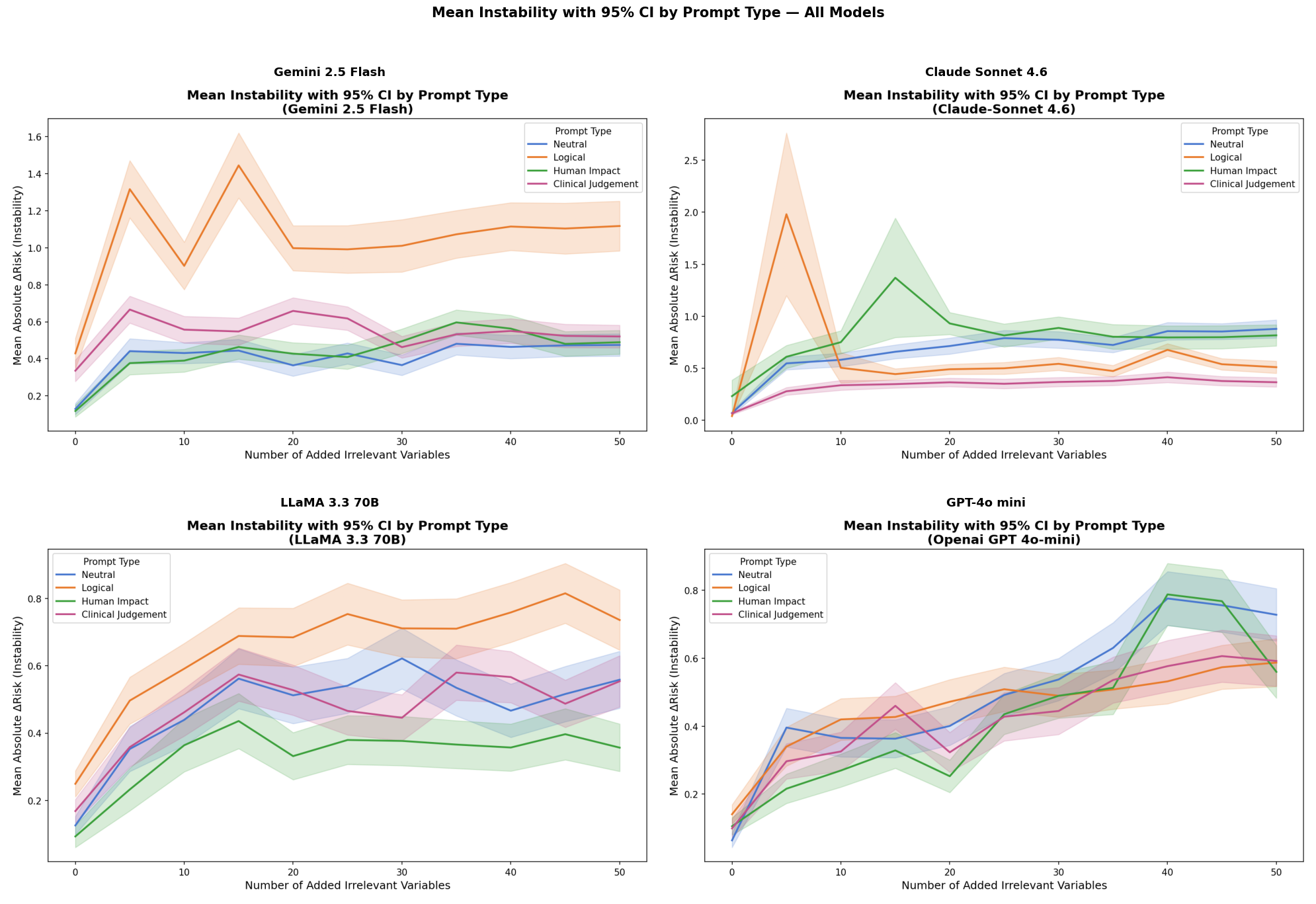}
    \caption{Mean instability with 95\% CI by prompt type. Instability rises most significantly at the first addition of insignificant variables and is on average higher at the logical prompt framing and lowest at the clinical judgment prompt framing. }
    \label{fig:means}
\end{figure}

\begin{figure}
    \centering
    \includegraphics[width=\linewidth]{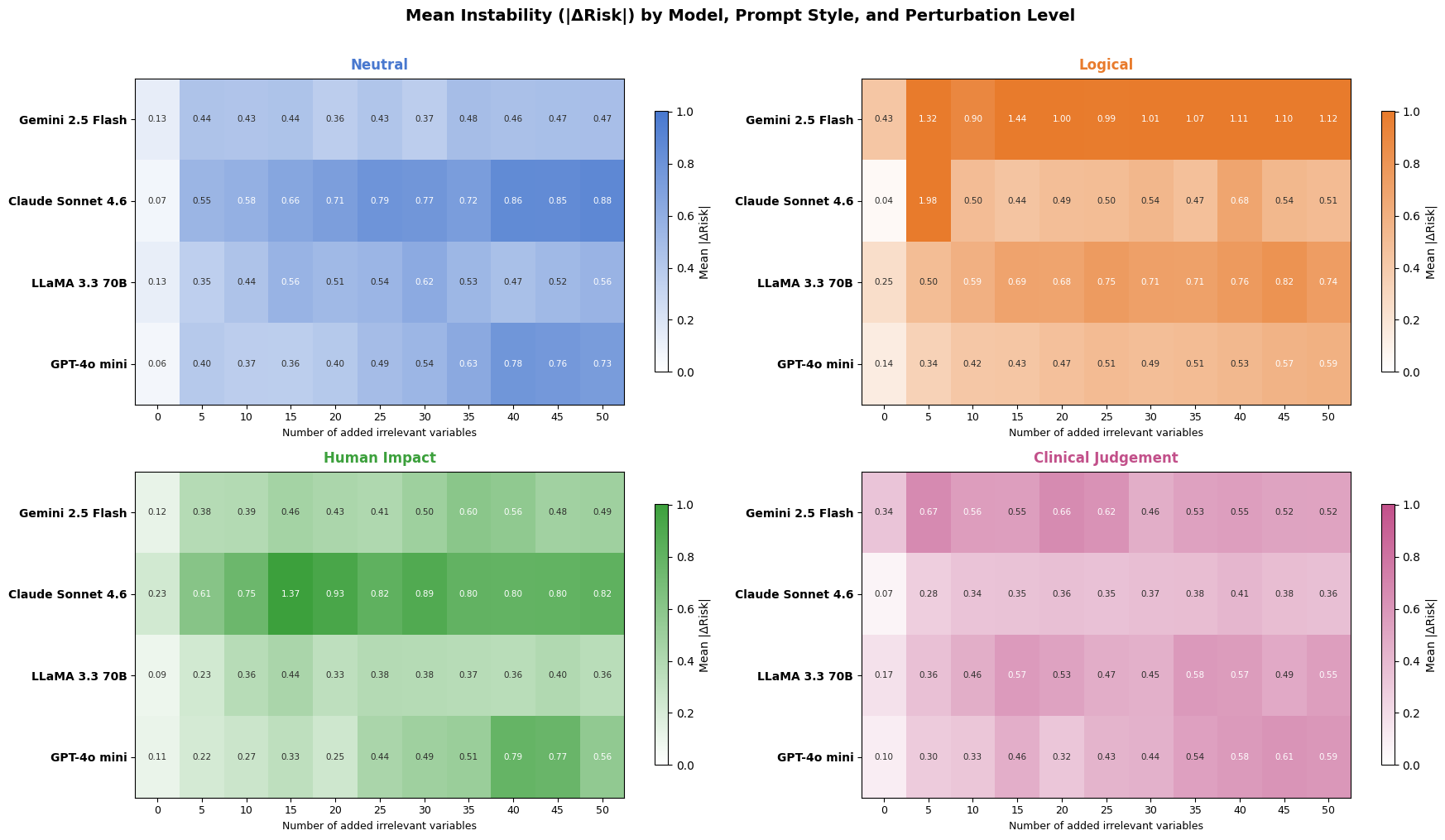}
    \caption{Mean Instability heatmap by model, prompt, and perturbation level}
    \label{fig:alt}
\end{figure}

Models also demonstrate an increase in mean risk score instability with the addition of clinically irrelevant variables, as shown by the Simple $\beta$1 term across the majority of model--prompt combinations (Figure~\ref{fig:means}). GPT-4o mini showed the most consistent results, rejecting the null across all prompt conditions ($p < 0.001$ throughout). For LLaMa, we reject the null under neutral, logical, and clinical judgment conditions ($p < 0.01$), but fail to reject under human impact ($p = 0.183$). Claude rejected the null under neutral ($p < 0.01$) and clinical judgment ($p = 0.012$), but failed to reject under human impact and logical prompting. Gemini failed to reject the null under Simple $\beta$1 for the majority of prompt conditions, with neutral ($p = 0.057$), logical ($p = 0.673$), and clinical judgment ($p = 0.954$) all non-significant, and only human impact reaching significance ($p =0.005$), suggesting that in Gemini added features alone does not reliably drive instability without considering prompt context. Across all models, the addition of the first five irrelevant variables produces the largest single shift in mean instability (Figure~\ref{fig:alt}), suggesting models are more sensitive to the initial introduction of irrelevant information, rather than the gradual increase.

Evaluation of the variable-level weight attributions reported in model responses reveals a pattern where models frequently assigned comparable numerical values to clinically insignificant characteristics as to clinically relevant indicators. In many responses, variables such as liking vegetables (-0.2), favorite sport(-0.2), and drinking coffee(-0.1), were given similar value to more relevant features like age, socioeconomic status, or in some cases even current diagnosis.

More notably, several of these attributions reflected implicit moral judgments towards patient behaviors that have no established relationship with clinical evaluation. Variables such as donating blood (-0.2, attributed to ``engagement in health-promoting activities"), picking up litter (-0.1, attributed to ``healthy lifestyle choices"), using public transportation (+0.1, attributed to ``potential for reduced mobility or social isolation"), having a drivers license(+0.5 attributed to ``potential barrier to accessing care or maintaining independence"), and being bilingual (-0.5 attributed to ``potential protective factor for cognitive reserve and social connection")
 were all assigned directional risk contributions with accompanying clinical-sounding rationales. The models appeared to reward behaviors it deemed prosocial or health conscious, although the magnitude and even direction of these values were not held consistent across patients. As insignificant variables were randomly introduced from the synthetic dataset, this pattern describes a lack of meaningful distinctions between clinical and non-clinical features in the models output. This is not only a statistical concern, but an ethical one as it suggests a form of algorithmically coded bias where patients may receive higher or lower risk scores based on behaviors that reflect preferences or practices rather than clinical necessity.

\section{Discussion} 

This study examined whether LLM-generated risk scores remain stable when clinically insignificant information is systematically introduced, and whether prompt formulation affects this stability. Across all four models and prompt styles, clinically insignificant features significantly destabilized model outputs, and prompt variations independently affected the trajectory of instability in a model-dependent manner, together questioning the reliability of LLMs as clinical decision support tools.

Instability is not only affected by insignificant features,  but also by how the clinical question is asked. The logical prompt was the only style associated with significant instability effects across all four models,  a level shift in Gemini (Complex $\beta_2$, 
$p < 0.01$) and LLaMA (Complex $\beta_2$, $p = 0.042$), and a 
significant interaction with noise accumulation in Claude (Complex $\beta_2$, 
$p < 0.01$) and GPT-4o mini (Complex $\beta_2$, $p = 0.001$), potentially due to its instruction to numerically weigh every feature. No single prompt was universally optimal across architectures, and the largest instability shift occurred at the first addition of irrelevant variables across all models (Figure~\ref{fig:alt}), with the rate of increase slowing thereafter in a model-dependent manner.

\paragraph{Clinical Implications}: These results signify that LLMs attributional inconsistencies raises fairness concerns in its assessment of patients. Those from different socioeconomic, cultural, or lifestyle backgrounds may receive systematically different risk scores for reasons unrelated to their clinical profile. Therefore, future LMM-based diagnostic systems should be evaluated for stability under more realistic patient information, including the presence of non-clinical contextual data while may be present in patient records in the real world. 

\paragraph{Technical Implications}: When utilizing LLMs for clinical decision making, prompt design should be treated as a controlled methodological variable rather than an afterthought in any pipeline. The instability introduced by prompt variations is arguably greater than instability introduced by and increase in insignificant variables in several models tested. The perturbation design also serves as a reusable framework that can be used in other LLM evaluations of clinical tasks. Additionally, the response format of outputs should be included as an evaluation metric, as certain prompt designs make it more difficult to extract data due to structurally unstable outputs. And finally, cross-model evaluations under varying prompts should be considered a minimum standard for any LLM-based clinical decision making study.

\paragraph{Limitations and Future Directions}

Reliability audits are difficult due to the ``black-box nature" of the model architectures. Although models were prompted to provide variable attribution values and a rationale, no mechanism was used to viably confirm that the explanations accurately reflect the attributions. It is possible that the variables a model claims to have weighted heavily did not meaningfully impact the score, and conversely, variables that were given little significance may have affected the output. This limits the interpretation of the qualitative findings and emphasizes the need for mechanistic verification in future works. It would be beneficial to employ a standardized attribution entropy metric to quantify qualitative findings. Measuring how evenly model-assigned weights are distributed across clinical and non-clinical features would allow the instability of variable attributions to be measured consistently across models and prompts.

The synthetic patient cohort poses another limitation. While conditional probabilities were induced to reflect realistic clinical diversity, the dataset does not capture the full complexity of real psychiatric patient records, including detailed clinical notes and comorbidity patterns. This dataset consisted of randomly selected insignificant variables and does not represent the full range of non-clinical information that may be present in a patient profile. The data also lacks any form of missing information, which is entirely viable in real clinical notes or patient profiles. The patient sample should also be further evaluated, examining whether the instability is evenly distributed among patient subgroups. Certain patients with specific demographics or clinical features may be more susceptible to instability in risk scores. For instance, patients from lower socioeconomic background or minority groups may get different scores if the non-relevant behavioral and lifestyle variables correlate with the presented demographics, thus the resulting instability may be attributed to confounding bias. If instability is shown to be not uniformly unstable across patient types, this may raise concerns in health equity when utilizing AI-assisted diagnostic systems.

Furthermore, as much of this information would be patient-reported, it is unlikely that all of the presented information would be accurate. Electronic health record (EHR) notes contain unstructured narratives, ambiguous language, and clinically embedded context that may interact with model reasoning in ways that structured feature lists do not capture. It is possible that instability patterns observed here would differ substantially when models are presented with EHR notes. Future works would emphasize the use of anonymous and unstructured EHR notes, and longitudinal patient history before establishing any generalizations to real world deployment of LLM diagnostic systems. 

% In general, these results support the approach as an auditing strategy. Accordingly, the replication of the perturbation methodology using unstructured EHR notes rather than synthetic tabular data would strengthen the validity of these findings in a clinical setting. Insignificant details embedded in textual notes would establish whether the instability persists when the information presented is less clearly defined. 

% The prompt sensitivity suggests that future works should examine whether fine-tuning models on clinician assessments reduce dependence on prompts, as the clinical judgment variation stabilized outputs for some models. Prompts could also be designed to focus solely on clinical features to potentially reduce the tendency to value insignificant information.

\subsection{Conclusions}
This study shows that LLM-generated risk scores are sensitive to clinically irrelevant information, and that this instability depends on both the model and the prompt. Across 44,000 simulations, adding insignificant features led to statistically significant shifts in outputs in nearly all settings. Clinical-style prompting reduced this effect somewhat, while logical prompting increased it. The results also show that models can assign meaning to non-clinical variables, reflecting implicit value judgments rather than evidence-based reasoning. Instability plateaus after only a few irrelevant features are added, indicating that even small amounts of extraneous information can meaningfully alter outputs. In practice, this means most real clinical records (where non-essential details are common) \textit{may already contain enough noise to make model behavior unreliable}. These findings indicate that model audits across prompt design and clinically insignificant features, such as the one presented, should be explicitly evaluated before deployment in clinical workflows. 
% ACKNOWLEDGEMENTS ONLY GO IN THE CAMERA-READY, NOT THE SUBMISSION
% \acks{Many thanks to all collaborators and funders!}

%Do NOT change font size of references or modify the bibliography style
\bibliography{references}

\newpage
\appendix
\section*{Appendix A.}
Our full list of clinically insignificant features are: Political Ideology, Eye Color, Likes Chocolate, Pet Preference, Recycling Habits, Drinks Coffee, Annual Charity Donation, Favourite Color, Favourite Season, Favourite Music Genre, Favourite Movie Genre, Favourite Cuisine, Orders Takeout, Spotify vs AppleMusic, Preferred Grocery Store, Favourite Super Store, Morning or Night, Birthday Month, Favourite Meal, Likes Art, Favourite Sport, Religious, Exercises, Likes Kids, Travels, Likes Vegetables, Phone, Has Drivers License, Can Ride Bicycle, Best Highschool Subject, Skips Breakfast, Bilingual, Binge Watches TV, Has Cheated on an Assignment Before, Relationship Status, Academic Preference, Reads Novels, Water Intake (oz drank daily), Hair Color, Height in Inches, Handedness, Wears Glasses, Allergy, Registered Organ Donor, Volunteers, Uses Public Transportation, Donated Blood, Picks Up Litter, Helps Elderly, Votes.
\end{document}